\documentclass[10pt,twocolumn,letterpaper]{article}

\usepackage[pagenumbers]{iccv}              

\definecolor{iccvblue}{rgb}{0.21,0.49,0.74}
\usepackage[pagebackref,breaklinks,colorlinks,allcolors=iccvblue]{hyperref}
\usepackage{amsmath}
\usepackage{amsfonts}
\usepackage[ruled,linesnumbered]{algorithm2e}
\usepackage{arydshln}
\usepackage{multirow}
\usepackage{xcolor}
\usepackage[accsupp]{axessibility}

\definecolor{darkred}{rgb}{0.5, 0.0, 0.0} 
\definecolor{darkblue}{rgb}{0.0, 0.5, 0.0}

\title{Outlier-Aware Post-Training Quantization for Image Super-Resolution}

\author{
Hailing Wang$^{*}$ \quad Jianglin Lu \quad Yitian Zhang \quad Yun Fu \\[0.2cm]
Northeastern University, USA\\
{\tt\small \{wang.haili, lu.jiang\}@northeastern.edu, markcheung9248@gmail.com, yunfu@ece.neu.edu}\\
}

\begin{document}
\maketitle
\renewcommand{\thefootnote}{\fnsymbol{footnote}}
\footnotetext[1]{Corresponding author.}
\begin{abstract}
Quantization techniques, including quantization-aware training (QAT) and post-training quantization (PTQ), have become essential for inference acceleration of image super-resolution (SR) networks. Compared to QAT, PTQ has garnered significant attention as it eliminates the need for ground truth and model retraining. However, existing PTQ methods for SR often fail to achieve satisfactory performance as they overlook the impact of outliers in activation.
Our empirical analysis reveals that these prevalent activation outliers are strongly correlated with image color information, and directly removing them leads to significant performance degradation.
Motivated by this, we propose a dual-region quantization strategy that partitions activations into an outlier region and a dense region, applying uniform quantization to each region independently to better balance bit-width allocation. 
Furthermore, we observe that different network layers exhibit varying sensitivities to quantization, leading to different levels of performance degradation.
To address this, we introduce sensitivity-aware finetuning that encourages the model to focus more on highly sensitive layers, further enhancing quantization performance.
Extensive experiments demonstrate that our method outperforms existing PTQ approaches across various SR networks and datasets, while achieving performance comparable to QAT methods in most scenarios with at least a 75 speedup.
\end{abstract}    
\section{Introduction}
\label{sec:intro}

The goal of image super-resolution (SR) is to enhance image resolution, often by factors of 4$\times$ or more, while preserving content and detail. 
Although deep learning-driven SR models have attained superior results \cite{dong2015image, zhang2018image, mei2023pyramid}, these advancements come at the cost of increased parameter counts.  
With the growing demand for deploying SR models on edge devices and handling larger input sizes, there is an increasing need for models that balance both parameter efficiency and computational speed. 
To mitigate this issue, a range of compression techniques has been studied, including distillation \cite{wang2022edge, zhang2023data, li2024knowledge}, pruning \cite{yu2023dipnet, zhanglightweight, xia2023structured}, quantization \cite{tu2023toward, tu2022adabin}, and efficient module design \cite{park2018efficient, zhang2022efficient}. 
In this paper, we focus on image SR quantization, which not only reduces memory consumption but also significantly improves inference speed.

\begin{figure}[t]
    \centering
    \begin{subfigure}[b]{0.236 \textwidth}
           \centering         \includegraphics[scale=0.19,trim=0 0 10 0,clip]{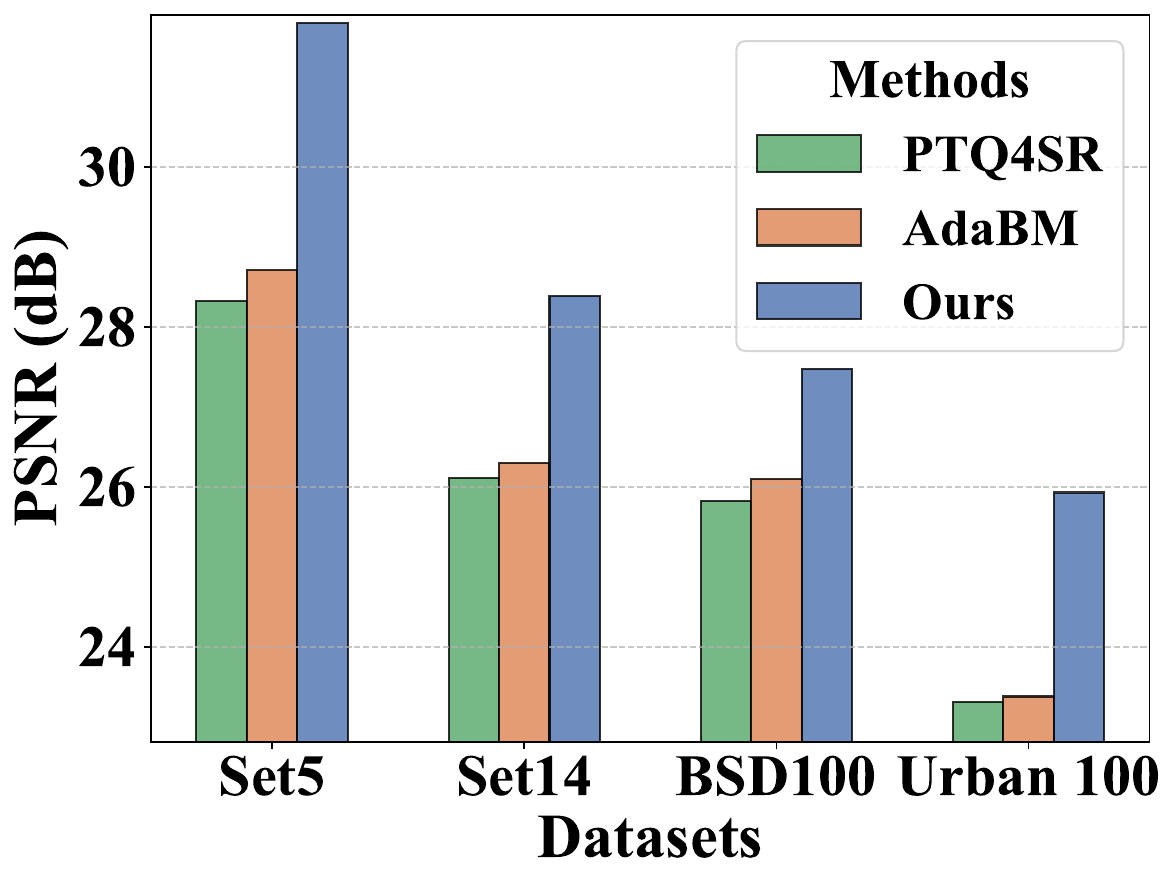}
            \caption{Comparison with PTQ methods in PSNR using RDN under W4A4.}
            \label{fig:mcl_psnr}
    \end{subfigure}
    \hfill
    \begin{subfigure}[b]{0.236 \textwidth}
            \centering       \includegraphics[scale=0.162,trim=3 0 0 0,clip]{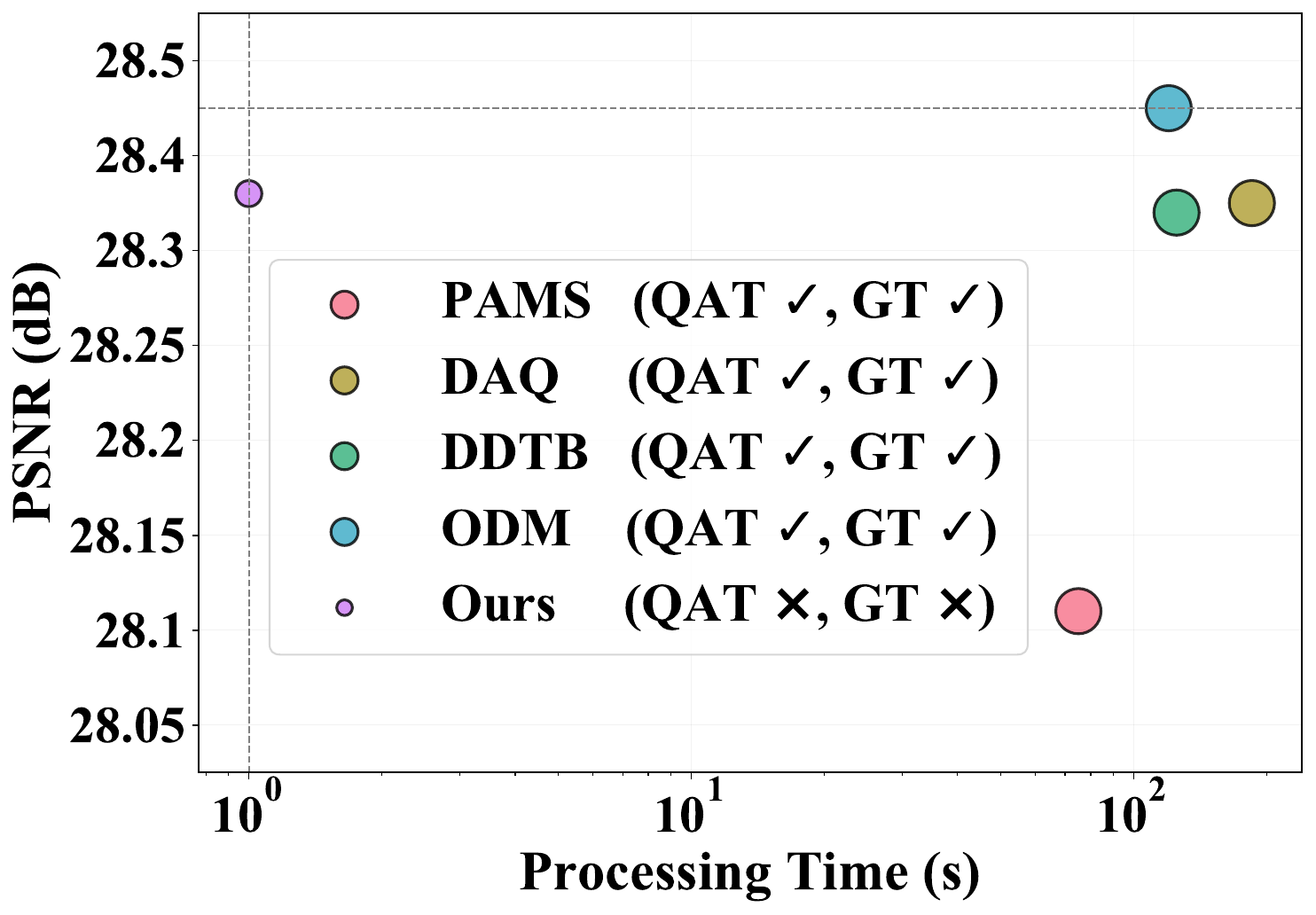}
            \caption{Comparison with QAT methods in PSNR using EDSR under W4A4.}
            \label{fig:mcl_ssim}
    \end{subfigure}
    \caption{Comparison of our method with SOTA PTQ and QAT baselines. In (b), GT denotes ground truth, the bubble size indicates the amount of training data required, and performance is averaged across four datasets.}
\label{fig:teaser}
\end{figure}

The goal of model quantization is to shrink the network’s parameters and activations (feature maps) from high precision to a compact representation while maintaining its original performance. Current quantization approaches are typically grouped into quantization-aware training (QAT) \cite{nagel2022overcoming, shen2021once, huang2023efficient} and post-training quantization (PTQ) \cite{liu2021post, shang2023post, yuan2022ptq4vit, fang2020post}, distinguished by whether they require retraining network weights.
QAT requires labeled data pairs and model retraining to adapt to the quantization process, whereas PTQ applies quantization without weight updates, making it a more source-efficient alternative to QAT.

In image SR, quantization has been predominantly explored through QAT \cite{li2020pams, hong2022daq, nagel2022overcoming, zhong2022dynamic}, with only a few PTQ methods \cite{tu2023toward, hong2024adabm} proposed. 
This is primarily because PTQ, particularly in activation quantization, suffers from greater performance degradation compared to QAT.
To mitigate this issue, Tu \textit{et al.} \cite{tu2023toward} introduce the first PTQ method for SR, employing a density-based double cropping technique to constrain the activation distribution within a manageable range. 
However, directly clipping activations outside the selected range may lead to significant performance degradation. 
Subsequently, Hong et al. \cite{hong2024adabm} propose a dynamic quantization technique that adjusts bit allocation based on input variations. This is motivated by the observation that assigning different bit widths to various input images and network layers improves quantization performance. While effective, ensuring hardware compatibility for dynamic quantization remains a challenge. 

In this paper, we first analyze the activation distributions of SR models and observe that activation outliers are prevalent across various networks (see Figure \ref{fig:feature_map_distribution}).
To investigate their impact, we compare the visual quality of images with and without outliers.
Our empirical findings reveal that outliers are strongly correlated with image color information, and directly removing them leads to noticeable color shifts and significant performance degradation (see Figure \ref{fig:clip99}). 
This underscores the importance of preserving outliers in quantization to maintain color fidelity and enhance overall quantization performance. 
However, retaining outliers using existing methods consumes a substantial portion of the bit width allocated for normal activations, severely compromising quantization effectiveness. 
To address this, 
we propose a dual-region quantization strategy that partitions activations into an outlier region and a dense region. 
We then apply uniform quantization to each region independently, ensuring a more balanced bit-width allocation between these two regions. Furthermore, we observe that different network layers exhibit varying sensitivity to quantization, as evidenced by the extent of performance degradation when each layer is quantized individually. Based on this insight, we propose a sensitivity-aware loss that encourages the model to focus more on highly sensitive layers, further enhancing overall quantization performance.

Figure \ref{fig:teaser} presents a comparison of our approach against state-of-the-art (SOTA) PTQ and QAT methods on representative SR networks. 
Figure \ref{fig:mcl_psnr} shows that our method consistently outperforms PTQ baselines across various datasets. 
From Figure \ref{fig:mcl_ssim}, our method achieves performance comparable to QAT methods, despite not requiring retraining or ground truth, while providing significantly higher efficiency.
To summarize, our core contributions are:

\begin{itemize}

    \item Our empirical analysis reveals that activation outliers are strongly correlated with color information, and removing them leads to significant color shifts in generated images. 
    
    \item We identify an allocation trade-off between outliers and normal activations: clipping outliers causes severe performance degradation, while retaining them consumes the bit width allocated for normal ones. To address this, we quantize outliers and normal activations separately, ensuring a more balanced and effective bit-width allocation.

    \item We uncover that different layers exhibit varying sensitivities to quantization and propose a sensitivity-aware loss function to focus more on highly sensitive layers.

    \item Comprehensive evaluations show that the proposed approach exceeds existing SOTA PTQ baselines and achieves performance comparable to QAT  methods, while delivering a $75$ $\times$ speedup. 
    
\end{itemize}

\section{Related Works}
\subsection{Efficient Image Super-Resolution}
Efficient SR models fall into several categories, including architectural design, neural architecture search (NAS), knowledge distillation (KD), pruning, and quantization. For efficient architectural design, Ahn \textit{et al.} \cite{ahn2018fast} introduce cascading residual connections and efficient residual blocks to construct a compact SR network. 
Sun \textit{et al.} \cite{sun2023spatially} propose an efficient feature modulation that combines CNN-like efficiency with transformer adaptability. 
Regarding NAS-based methods, Huang \textit{et al.} \cite{huang2022differentiable} present a differentiable NAS strategy to identify efficient SR networks, integrating both unit-level and network-level search spaces to optimize SR quality. For KD-based approaches, Zhang \textit{et al.} \cite{zhang2021data} introduce a novel data-free knowledge distillation framework for SR, which is adaptable to various teacher-student configurations. 
In pruning-based solutions, Wang \textit{et al.} \cite{wang2021exploring} develop a SR network with sparse masks that simultaneously exploit spatial and channel dimensions to jointly identify and remove unnecessary computation at a fine-grained level. In this paper, we focus on quantization-based methods for image SR, as they effectively reduce memory consumption while significantly improving inference speed.

\subsection{Quantization for Image Super-Resolution}
Quantization methods, including QAT \cite{jiang2021training, li2020pams, wang2021fully, hong2022daq, zhong2022dynamic, hong2022cadyq, wei2023ebsr, qin2024quantsr} and PTQ \cite{tu2023toward, hong2024adabm}, have both been explored for image SR. 
The first QAT-based SR work \cite{li2020pams} introduces a trainable truncation parameter to adaptively constrain the quantization range, motivated by the observation that SR models without batch normalization typically exhibit a large dynamic range. Wang \textit{et al.} \cite{wang2021fully} propose a quantizer with learnable margins, enabling adaptability to variation in weights and activations from one layer to another. To further address dynamic range issues, Hong \textit{et al.} \cite{hong2022daq} develop a channel distribution-aware quantization scheme.
Additionally, some approaches \cite{hong2022cadyq, tian2023cabm} employ dynamic quantization strategies with adaptive bit-width allocation for different inputs and layers. 
However, ensuring hardware compatibility for dynamic quantization remains an open challenge.
In contrast, PTQ for image SR has received significantly less attention. 
The first PTQ-based SR method \cite{tu2023toward} introduces a density-based dual clipping and pixel-aware calibration to optimize the quantization parameters. 
Subsequently, Hong \textit{et al.} \cite{hong2024adabm} introduce a dynamic quantization method with adaptive bit mapping. 
While effective to some extent, these methods largely overlook the impact of outliers, resulting in suboptimal quantization performance.
In this work, we emphasize the importance of outliers in quantization, revealing that  outliers are strongly correlated with image color information.

\begin{figure*}[!tb]
    \centering
    \includegraphics[width=0.95\textwidth]{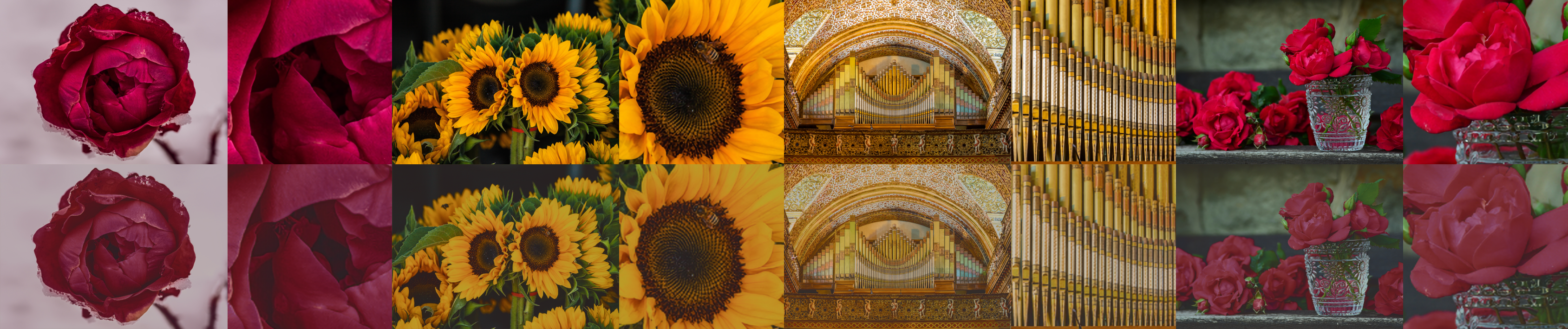}
    \caption{After clipping $1\%$ of activation outliers in the full-precision model, the outputs (bottom) exhibit noticeable color distortions compared to the original ones (top), affecting both global regions and detail-rich local areas of the images.}
    \label{fig:clip99}
\end{figure*}

\section{Methodology}

\subsection{Preliminaries}
Quantization reduces parameter precision, thereby shrinking memory footprints and accelerating inference.
The process of quantizing a floating-point tensor \( x \) into a \( b \)-bit unsigned integer can be formally described as follows: 
\begin{equation}
x_{\text{int}} = \left\lfloor \frac{\mathit{clamp}(x, l, u) - l}{u - l} \times (2^b - 1) \right\rceil,
\end{equation}
where \( l \) and \( u \)  are the lower and upper bounds of \( x \) respectively, \( \mathit{clamp}(x, l, u) = \min(\max(x, l), u) \) restricts \( x \) within the bounds \( l \) and \( u \), and the function $\left\lfloor \cdot \right\rceil$ outputs the nearest integer of the input.
The quantized floating-point value \( x_q \) can be reconstructed from $x_{\text{int}}$ within the integer space to approximate the original value \( x \) via:
\begin{equation}
x_q = x_{\text{int}} \cdot \frac{u - l}{2^b - 1} + l.
\end{equation}
When \( x \) exhibits a (approximately) symmetric distribution around zero, the bounds \( u \) and \( l \) can be set to symmetric limits, such as \( u \) and \( -u \). This adjustment simplifies the quantization process by ensuring symmetry around zero.

Due to the highly asymmetric distribution of activations in SR networks \cite{hong2022cadyq, hong2022daq,tian2023cabm,zhong2022dynamic} and the relatively low sensitivity of weights to quantization \cite{tu2023toward}, an asymmetric quantizer is typically applied to activations, whereas a symmetric uniform quantizer is utilized for weights.
Compared to weight quantization, previous studies \cite{hong2022daq, tu2023toward} have shown that activation quantization is the primary cause of performance degradation in SR models.
Therefore, this paper also focuses on activation quantization of SR models.

\subsection{Observation \& Motivation}

\begin{figure}[!tb]
    \centering
    \includegraphics[scale=0.34,trim=0 10px 0 0,clip]{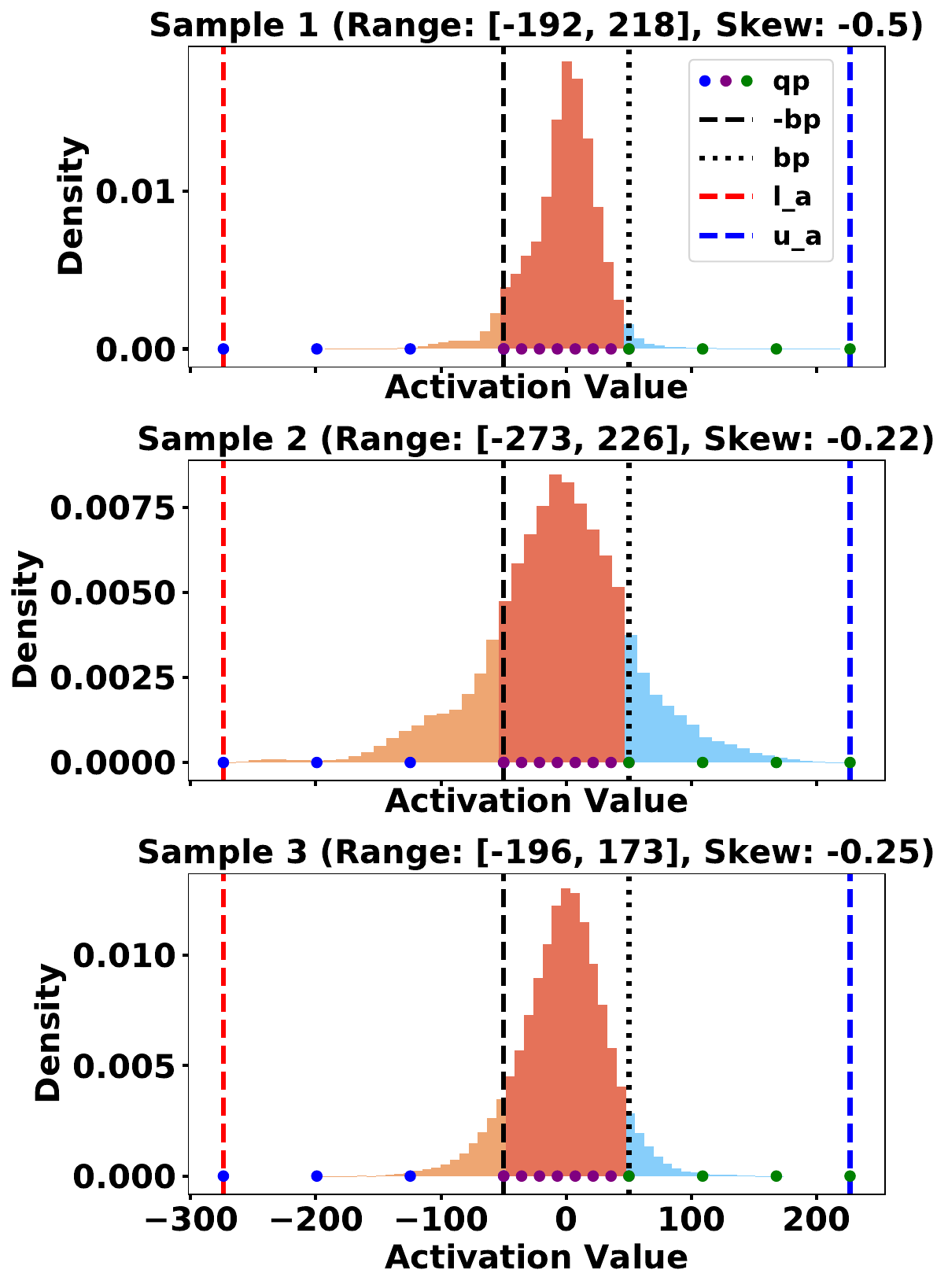}

    \caption{The activation distributions of three different samples at the same layer (\texttt{body.15.conv1}) in EDSR exhibit variations in range (\texttt{Range}) and skewness (\texttt{Skew}). These distributions are divided by a breakpoint $bp$ into a dense region \([-bp, bp]\) and an outlier region \([l\_a, -bp) \cup (bp, u\_a]\), both of which undergo uniform quantization to the corresponding quantization points $qp$.}
    \label{fig:feature_map_distribution}
\end{figure}

\textbf{Observation 1}. 
In Figure \ref{fig:feature_map_distribution}, we illustrate the activation distribution of different samples at the same layer (\texttt{body.15.conv1}) of the EDSR network. 
It is evident that all samples contain outliers in their distributions. Most activation values are concentrated within a shallow range (e.g., $[-50, 50]$), which we refer to as the \textit{dense region}. Outliers, on the other hand, are located beyond this region, forming what we call the \textit{outlier region}.
Notably, the bounds of the outlier region vary significantly across different samples. For instance, the left bound for sample $1$ is $-192$, whereas for sample $2$, it is $-273$.
From this observation, two key questions arise: {{Do these outliers influence the quality of the restored images, and what specific features do they correspond to in the generated images?}}

To answer these questions, we clip the outliers ($1\%$ of activations) in the feature map and visualize the resulting images in Figure \ref{fig:clip99}. 
The visual comparison clearly shows that removing outliers leads to noticeable color distortion in both global and local regions of the images, such as faded flower colors. 
\textit{This observation indicates that activation outliers are closely linked to image color information and should be preserved during the quantization process.} 
Based on this insight,  we propose outlier-aware quantization to minimize quantization errors in Section \ref{colorinfo}.

\begin{figure}[!tb]
    \centering
    \includegraphics[width=0.46\textwidth]{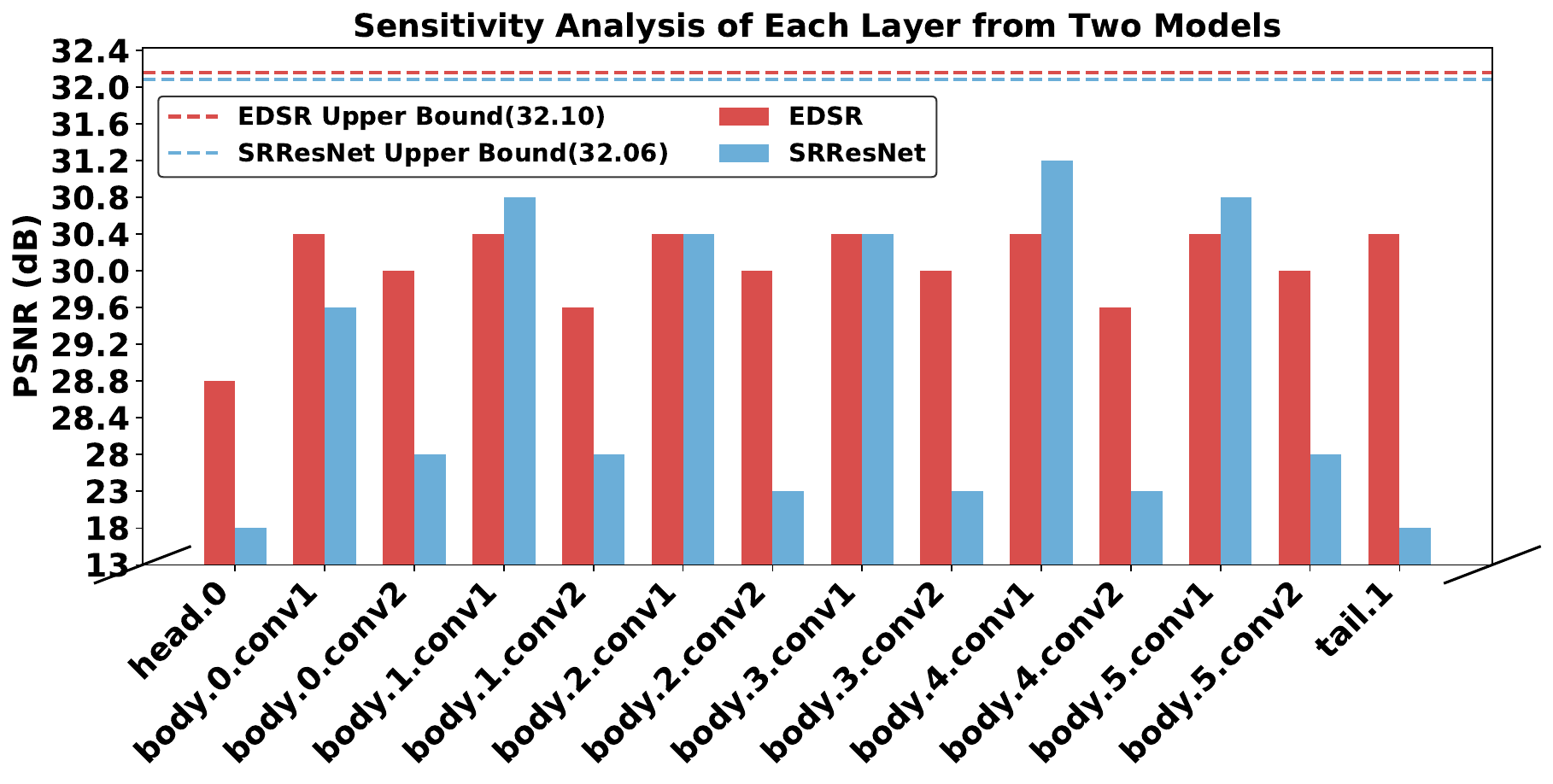}
    \caption{Performance comparison of 4-bit quantization applied individually to each layer of EDSR and SRResNet. Certain layers show a notable drop compared to the upper bound (full-precision performance), indicating higher sensitivity to quantization.}    
    \label{fig:activation_sensitivity}
\end{figure}

\textbf{Observation 2}. 
We further analyze quantization error across different layers by independently quantizing activations in each layer of the EDSR and SRResNet networks.
To evaluate quantization performance, we compute the average PSNR between the quantized images and the ground truth (GT) across $100$ randomly selected image pairs from the DIV2K dataset \cite{agustsson2017ntire}.
As shown in Figure \ref{fig:activation_sensitivity}, different network layers exhibit varying degrees of sensitivity to quantization.
While some layers experience significant performance degradation, others remain robust to quantization.
For instance, in SRResNet, the \texttt{head.0} layer suffers a substantial drop in PSNR, plunging from $32.06$ dB to $18.26$ dB. In contrast, certain layers, such as \texttt{body.4.conv1}, maintain high performance with PSNR values of up to $31.20$ dB.
\textit{Motivated by this observation, we propose  focusing more attention on highly sensitive layers in quantization rather than distributing equal attention across all layers.} 
To achieve this, we introduce a sensitivity-aware loss in Section \ref{sensitivityaware}.
\subsection{Piecewise Linear Quantizer}
\label{colorinfo}
As demonstrated in Observation 1, preserving activation outliers is crucial for retaining image information. 
However, retaining outliers during quantization will consume a substantial portion of the bit width allocated for normal activations, reducing the representation space available for them. Inspired by \cite{fang2020post}, to achieve a balanced bit-width allocation between outliers and normal activations, we propose a dual-region quantization strategy that partitions activations into two distinct, non-overlapping regions and designs piecewise linear quantization to quantize  each region independently.
This method preserves the unique characteristics of both normal activations and outliers.

Specifically, given an asymmetric activation range \(R=[l_a, u_a]\), where $l_a$ and $u_a$ denote the lower and upper activation bounds, we introduce a learnable breakpoint \( bp \) to divide the range \(R\) into a symmetric dense region \( R_1 = [-bp, bp] \), which contains most normal activations, and an outlier region \( R_2 = R_2^- \cup R_2^+ = [l_a, -bp) \cup (bp, u_a] \), which captures the extreme values in the activation distribution.  Our piecewise linear quantizer converts a floating-point tensor \( x \) into a \( b \)-bit integer representation via:
\begin{equation}
\scalebox{0.85}{$
x_{\text{int}} \leftrightarrow 
\begin{cases}
    \left\lfloor \frac{\mathit{clamp}(x, -bp, bp)}{2bp} \times \left(2^{b-1} - 1\right) \right\rceil, & x \in R_1 \\[8pt]
    \left\lfloor \frac{\mathit{clamp}(x, l_a, -bp) - l_a}{-bp - l_a} \times \left(2^{b-2} - 1\right) \right\rceil, & x \in R_2^- \\[8pt]
    \left\lfloor \frac{\mathit{clamp}(x, bp, u_a) - bp}{u_a - bp} \times \left(2^{b-2} - 1\right) \right\rceil, & x \in R_2^+
\end{cases}
$}
\end{equation}
where values in both regions are quantized at the same bit level.
We determine appropriate values for \( l_a \), \( u_a \), and \( bp \) through a statistical analysis of a calibration set. 
Specifically, for the first batch, we initialize \( l_a \) as the minimum activation value, \( u_a \) as the maximum activation value, and \( bp \) as the $99$th percentile of the activation values. 
For weight quantization, we set the upper bound of weights \( u_w \) as the maximum absolute value in each layer. 
In subsequent batches, these parameters are updated using an exponential moving average \cite{finch2009incremental}.

\subsection{Sensitivity-Aware Finetuning}
\label{sensitivityaware}
While static statistical analysis on calibration data provides initial estimates for parameters \(l_a\), \(u_a\), \(u_w\), and \(bp\) across different layers, the substantial variation in outlier ranges across samples within the same layer may affect the accuracy of these estimates.
To address this, we refine these quantization parameters by finetuning the model on the calibration data.
Inspired by Observation 2, which highlights that different layers exhibit varying sensitivities to quantization, we design a layer-specific loss function and perform sensitivity-aware finetuning.
This strategy directs the model to focus more on highly sensitive layers during  quantization rather than distributing equal attention across all layers, thus enhancing the model's adaptability to the dynamic nature of activation distributions across different samples.

To quantify the sensitivity of each layer to quantization, we pass the calibration data $\mathcal{D}_{\text{cal}}$ through a full-precision SR network $\mathcal{K}$ and compute the mean variance of the feature maps. 
We use this variance as an indicator of quantization sensitivity, where higher variance corresponds to greater sensitivity. 
The layer-wise quantization sensitivity \( s_{k} \) is defined as:
\begin{equation} \label{eq:sen}
s_k = \frac{\exp\left(\frac{1}{N} \sum_{x \in D_{\text{cal}}} \sigma(x_k)\right)}{\sum_{j=1}^{K} \exp\left(\frac{1}{N} \sum_{x \in D_{\text{cal}}} \sigma(x_j)\right)},
\end{equation}
where \( \sigma(x_k) \) represents the standard deviation of the feature map \(x_k\) at the \( k \)-th layer, \( N \) is the total number of batches in the calibration dataset, and \( K \) is the total number of layers in the SR network.
We optimize the quantization parameters using a training loss $L_{\text{all}}$, which consists of a reconstruction loss $\mathcal{L}_{\text{rec}}$ and a sensitivity-aware loss $\mathcal{L}_{\text{sen}}$, formulated as:
\begin{align}
\mathcal{L}_{\text{rec}} &= \frac{1}{N} \sum_{i=1}^{N} \left\| \mathcal{K}(I_{lr}^i) - \mathcal{Q}(I_{lr}^i) \right\|_{1}, \\
\mathcal{L}_{\text{sen}} &= \frac{s_k}{K} \sum_{k=1}^{K} \left\| \frac{F_{\mathcal{K}}^{k}}{\| F_{\mathcal{K}}^{k} \|_{2}} - \frac{F_{\mathcal{Q}}^{k}}{\| F_{\mathcal{Q}}^{k} \|_{2}} \right\|_{2}, \\ 
L_{\text{all}} &= \mathcal{L}_{\text{sen}} + \lambda \mathcal{L}_{\text{rec}},
\label{eq:all_loss}
\end{align}
where \(\lambda\) is a balancing parameter, \(I_{lr}^i\) denotes the $i$-th low-resolution input image, \(\mathcal{K}\) and \(\mathcal{Q}\) represent the pre-trained full-precision and quantized networks, respectively. 
\( F_{\mathcal{K}}^{k} \) and \( F_{\mathcal{Q}}^{k} \) denote the feature outputs at the \(k\)-th layer of \(\mathcal{K}\) and \(\mathcal{Q}\), respectively.
Notably, our approach requires only low-resolution images for computing \( L_{\text{all}} \), eliminating the need for ground-truth high-resolution images. This further enhances the practicality of our method.

During the fine-tuning phase, we update the quantization parameters in a staged manner for progressive optimization. 
Specifically,
in the first epoch, we update \( u_w \) while keeping all other parameters fixed. 
In the next epoch, only \( l_a \) and \( u_a \) are updated, keeping the rest unchanged.
And in the subsequent epoch, we update \( bp \) while keeping other parameters frozen. 
This cycle is repeated over multiple iterations to gradually refine the quantization parameters.
The overall quantization process is summarized in Algorithm \ref{algorithm1}.

\begin{algorithm}[h]
\caption{Quantization Algorithm}\label{algorithm1}
\KwIn{Full-precision SR network $\mathcal{K}$ with $K$ layers, calibration dataset $\mathcal{D}_{\text{cal}} = \{I_{lr}^i\}_{i=1}^N$, where $N$ is the number of calibration batches}
\KwOut{Quantized network $\mathcal{Q}$}

\textbf{Calibration Phase:} \\
\For{$i = 1, \dots, N$}{
    \For{$k = 1, \dots, K$}{
        \uIf{$i = 1$}{
            $u_w^k \gets \max | W^k |$,\\ 
            $l_{a,1}^k, u_{a,1}^k \gets \min(F^k(i)), \max(F^k(i))$,\\ 
            $bp_1^k \gets \text{Perc}_{99}(F^k(i))$\;
        }
        \Else{
            $l_{a,i}^k \gets \beta \cdot l_{a,i-1}^k + (1 - \beta) \cdot \min(F^k(i))$,\\
            $u_{a,i}^k \gets \beta \cdot u_{a,i-1}^k + (1 - \beta) \cdot \max(F^k(i)),$\\
            $bp_i^k \gets \beta \cdot bp_{i-1}^k + (1 - \beta) \cdot \text{Perc}_{99}(F^k(i));$\\
        }
    }
    Obtain $\{s_k^i\}_{k=1}^{K}$ using Eq.~(\ref{eq:sen})\;
}

\textbf{Fine-tuning Phase:} \;
\For{epoch = 1, \dots, \text{\#epochs}}{
    \uIf{$epoch \bmod 3 = 1$}{
        Update $\{u_w^k\}_{k=1}^{K}$ with Eq.~(\ref{eq:all_loss})\;
    }
    \uElseIf{$epoch \bmod 3 = 2$}{
        Update $\{l_a^k, u_a^k\}_{k=1}^{K}$ with Eq.~(\ref{eq:all_loss})\;
    }
    \Else{
        Update $\{bp^k\}_{k=1}^{K}$ with Eq.~(\ref{eq:all_loss})\;
    }
}
\end{algorithm}

\section{Experiments}
\begin{table*}[t]
    \centering
    \scalebox{0.95}{
    \resizebox{\textwidth}{!}{
    \begin{tabular}{lccccccccccccccc}
        \toprule
        \multirow{2}*{Method} &
        \multirow{2}*{FT} &
        \multirow{2}*{W / A} &
        \multicolumn{3}{c}{Set5} & 
        \multicolumn{3}{c}{Set14} & 
        \multicolumn{3}{c}{BSD100} & 
        \multicolumn{3}{c}{Urban100} \\
        \cmidrule(lr){4-6}\cmidrule(lr){7-9}\cmidrule(lr){10-12}\cmidrule(lr){13-15}
        & & & FAB & PSNR & SSIM & FAB & PSNR & SSIM & FAB & PSNR & SSIM & FAB& PSNR & SSIM \\
        \hline
        EDSR \cite{lim2017enhanced} & $-$ & 32 / 32 & 32.0 & 32.10 & 0.894 & 32.0 & 28.58 & 0.781 & 32.0 & 27.56 & 0.736 & 32.0 & 26.04 & 0.785 \\
        \hline
        EDSR-MSE \cite{choi2018bridging} & $\times$ & 6 / 6 & 6.0 & 31.84 & 0.887 & 6.0 & 28.37 & 0.775 &  6.0 & 27.45 & 0.731 &  6.0 & 25.73 & 0.775 \\
        EDSR-MinMax \cite{jacob2018quantization} & $\times$ & 6 / 6 &  6.0 & 31.56 & 0.866 &  6.0 & 28.26 & 0.760 &  6.0 & 27.29 & 0.714 &  6.0 & 25.76  & 0.760 \\
        EDSR-Percentile \cite{li2019fully} & $\times$ & 6 / 6 &  6.0 & 24.30 & 0.793 &  6.0 & 24.31 & 0.728 &  6.0 & 24.68 &  0.700 &  6.0 & 21.93 & 0.696 \\
        EDSR-PTQ4SR \cite{tu2023toward} & $\checkmark$ & 6 / 6 &  6.0 & 31.80 & 0.884 &  6.0 & 28.34 & 0.768 &  6.0 & 27.37 & 0.722 &  6.0 & 25.79 & 0.769 \\
        EDSR-AdaBM \cite{hong2024adabm} & $\checkmark$ & 6 / 6 &  5.7 & \textcolor{darkblue}{31.92} & \textcolor{darkblue}{0.887} &  5.6 & \textcolor{darkblue}{28.47} & \textcolor{darkblue}{0.777} &  5.4 & \textcolor{darkblue}{27.47} & \textcolor{darkblue}{0.731} &  5.7 & \textcolor{darkblue}{25.89} & \textcolor{darkblue}{0.778} \\
        EDSR-Ours & $\checkmark$ & 6 / 6 &  6.0 & \textcolor{darkred}{32.03} & \textcolor{darkred}{0.891} &  6.0 & \textcolor{darkred}{28.55} & \textcolor{darkred}{0.780} &  6.0 & \textcolor{darkred}{27.54} & \textcolor{darkred}{0.735} &  6.0 & \textcolor{darkred}{25.99} & \textcolor{darkred}{0.782} \\
        \hdashline
        EDSR-MSE \cite{choi2018bridging} & $\times$ & 4 / 4 & 4.0 & 27.74 & 0.827 &  4.0 & 26.03 & 0.734 &  4.0 & 25.95 & \textcolor{darkblue}{0.702} &  4.0 & 23.63 & 0.712 \\
        EDSR-MinMax \cite{jacob2018quantization} & $\times$ & 4 / 4  & 4.0 & 26.83 & 0.624  & 4.0 & 25.04 & 0.546  & 4.0 & 24.57 & 0.503  & 4.0 & 23.12 & 0.536 \\
        EDSR-Percentile \cite{li2019fully} & $\times$ & 4 / 4  & 4.0 & 24.03 & 0.776  & 4.0 & 23.95 & 0.712  & 4.0 & 24.42 & 0.687  & 4.0 & 21.62 & 0.677 \\
        EDSR-PTQ4SR \cite{tu2023toward} & $\checkmark$ & 4 / 4  & 4.0 & 30.51 & 0.836  & 4.0 & 27.62 & 0.735  & 4.0 & 26.88 & 0.693  & 4.0 & 24.92 & 0.721 \\
        EDSR-AdaBM \cite{hong2024adabm} & $\checkmark$ & 4 / 4  & 3.8 & \textcolor{darkblue}{31.02} & \textcolor{darkblue}{0.860} & 3.7 & \textcolor{darkblue}{27.87} & \textcolor{darkblue}{0.751} & 3.5 & \textcolor{darkblue}{26.91} & {0.700} & 3.7 & \textcolor{darkblue}{25.11} & \textcolor{darkblue}{0.736} \\
        EDSR-Ours & $\checkmark$ & 4 / 4 & 4.0 & \textcolor{darkred}{31.54} & \textcolor{darkred}{0.879} & 4.0 & \textcolor{darkred}{28.26} & \textcolor{darkred}{0.769} &  4.0 & \textcolor{darkred}{27.36} & \textcolor{darkred}{0.726} &  4.0 & \textcolor{darkred}{25.61} & \textcolor{darkred}{0.765} \\
        \hline
        RDN \cite{zhang2018residual} & $\times$ & 32 / 32 & 32.0 & 32.24 & 0.895 & 32.0 & 28.67 & 0.784 & 32.0 &  27.63 & 0.739 & 32.0 & 26.29 & 0.793 \\
        \hline
        RDN-MSE \cite{choi2018bridging} & $\times$ & 6 / 6 & 6.0 & 31.02 & 0.879 & 6.0 & 27.77 & 0.767 & 6.0 & 27.01 & \textcolor{darkblue}{0.724} & 6.0 & 25.01 & \textcolor{darkblue}{0.757} \\
        RDN-MinMax \cite{jacob2018quantization} & $\times$ & 6 / 6 & 6.0 & 30.59 & 0.863 &6.0 &  27.54 & 0.752 & 6.0 & 26.65 & 0.703 & 6.0 & 24.79 & 0.733 \\
        RDN-Percentile \cite{li2019fully} & $\times$ & 6 / 6 & 6.0 & 18.87 & 0.778 & 6.0 & 18.33 & 0.667 & 6.0 & 19.88 & 0.651 & 6.0 & 16.81 & 0.632 \\
        RDN-PTQ4SR \cite{tu2023toward} & $\checkmark$ & 6 / 6 & 6.0 & 30.73 & 0.877 & 6.0 & 27.60 & 0.765 & 6.0 & 26.85 & 0.720 & 6.0 & 25.08 & 0.756 \\
        RDN-AdaBM \cite{hong2024adabm} & $\checkmark$ & 6 / 6 & 5.7 & \textcolor{darkblue}{31.56} & \textcolor{darkblue}{0.881} & 5.6 & \textcolor{darkblue}{28.14} & \textcolor{darkblue}{0.769} & 5.5 & \textcolor{darkblue}{27.20} & {0.722} & 5.7 & \textcolor{darkblue}{25.31} & {0.755} \\
        RDN-Ours & $\checkmark$ & 6 / 6 & 6.0 & \textcolor{darkred}{32.20} & \textcolor{darkred}{0.894} & 6.0 & \textcolor{darkred}{28.62} & \textcolor{darkred}{0.782} & 6.0 & \textcolor{darkred}{27.61} & \textcolor{darkred}{0.738} & 6.0 & \textcolor{darkred}{26.24} & \textcolor{darkred}{0.790} \\
        \hdashline
        RDN-MSE \cite{choi2018bridging} & $\times$ & 4 / 4 & 4.0 & 25.55 & \textcolor{darkblue}{0.831} &  4.0 & 24.33 & \textcolor{darkblue}{0.725} &  4.0 & 24.49 & \textcolor{darkblue}{0.689} & 4.0 & 21.75 & \textcolor{darkblue}{0.692} \\
        RDN-MinMax \cite{jacob2018quantization} & $\times$ & 4 / 4 &  4.0 & 25.91 & 0.632 &   4.0 & 24.22 & 0.549 &  4.0 & 24.29 & 0.530 &  4.0 & 22.24 & 0.523 \\
        RDN-Percentile \cite{li2019fully} & $\times$ & 4 / 4 & 4.0 &  18.83 & 0.771 &  4.0 & 18.28 & 0.662 &  4.0 & 19.83 & 0.646 &  4.0 & 16.77 & 0.625 \\
        RDN-PTQ4SR \cite{tu2023toward} & $\checkmark$ & 4 / 4 &  4.0 & 28.32 & 0.813 &  4.0 & 26.11 & 0.709 &  4.0 & 25.82 & 0.671 & 4.0 &  23.31 & 0.668 \\
        RDN-AdaBM \cite{hong2024adabm} & $\checkmark$ & 4 / 4 &  3.8 & \textcolor{darkblue}{28.71} & {0.808} &  3.7 & \textcolor{darkblue}{26.30} & {0.707} & 3.6 & \textcolor{darkblue}{26.10} & {0.672} & 3.8 & \textcolor{darkblue}{23.38} & {0.663} \\
        RDN-Ours & $\checkmark$ & 4 / 4 &  4.0 & \textcolor{darkred}{31.80} & \textcolor{darkred}{0.885} &  4.0 & \textcolor{darkred}{28.39} & \textcolor{darkred}{0.775} & 4.0 &  \textcolor{darkred}{27.47} & \textcolor{darkred}{0.732} &  4.0 & \textcolor{darkred}{25.93} & \textcolor{darkred}{0.778} \\
        \bottomrule
    \end{tabular}
    }
    }
    \caption{Performance comparison of PTQ methods on W4A4 (4-bit weight and 4-bit activation) and W6A6 using EDSR and RDN as SR models, both with a scale factor of 4. 
    We add an additional FAB (Feature Average Bit-width) column specifically for the AdaBM method, as it is an adaptive PTQ method. 
    \textcolor{darkred}{Red} marks the highest quantization performance, while \textcolor{darkblue}{green} denotes the runner-up.}
    \label{tab:ptq}
\end{table*}

\begin{table}[t!]
    \centering
    \scalebox{0.48}{
    \resizebox{\textwidth}{!}{
    \begin{tabular}{lccccc}
        \toprule
        \raisebox{-1.5ex}{Method} & 
        \raisebox{-1.5ex}{W / A} & 
        \multicolumn{2}{c}{Test2K} & 
        \multicolumn{2}{c}{Test4K} \\
        & & PSNR & SSIM & PSNR & SSIM\\
        \hline
        EDSR \cite{lim2017enhanced}& 32 / 32 & 27.71 & 0.782 & 28.80 & 0.814 \\
        \hline
        EDSR-PTQ4SR \cite{tu2023toward} & 8 / 6  & 27.54 & 0.768 & \textcolor{darkblue}{28.91} & \textcolor{darkblue}{0.814}\\
        EDSR-AdaBM \cite{hong2024adabm} & 8 / 6  & \textcolor{darkred}{27.65} & \textcolor{darkred}{0.779} & {28.71} & {0.809}\\
        EDSR-Ours & 8 / 6  & \textcolor{darkblue}{27.59} & \textcolor{darkblue}{0.773} & \textcolor{darkred}{28.95} & \textcolor{darkred}{0.819} \\
        \hdashline
        EDSR-PTQ4SR \cite{tu2023toward} & 4 / 4 & 26.94 & 0.723 & 28.13 & 0.767\\
        EDSR-AdaBM \cite{hong2024adabm} & 4 / 4 & \textcolor{darkblue}{27.40} & \textcolor{darkblue}{0.758} & \textcolor{darkblue}{28.39} & \textcolor{darkblue}{0.784}\\
        EDSR-Ours & 4 / 4 & \textcolor{darkred}{27.49} & \textcolor{darkred}{0.767} & \textcolor{darkred}{28.83} & \textcolor{darkred}{0.814}\\
        \hline
        SRResNet \cite{ledig2017photo} & 32 / 32 & 27.64 & 0.781 & 28.72 & 0.813 \\
        \hline
        SRResNet-PTQ4SR \cite{tu2023toward} & 8 / 6 & 27.46 & 0.767 & \textcolor{darkblue}{28.78} & \textcolor{darkblue}{0.816}\\
        SRResNet-AdaBM \cite{hong2024adabm} & 8 / 6 & \textcolor{darkblue}{27.55} & \textcolor{darkred}{0.777} & 28.62 & 0.809 \\
        SRResNet-Ours & 8 / 6 & \textcolor{darkred}{27.55} & \textcolor{darkblue}{0.771} & \textcolor{darkred}{28.84} & \textcolor{darkred}{0.818} \\
        \hdashline
        SRResNet-PTQ4SR \cite{tu2023toward} & 4 / 4 & 27.06 & 0.749 & \textcolor{darkblue}{28.32} & \textcolor{darkblue}{0.797} \\
        SRResNet-AdaBM \cite{hong2024adabm}  & 4 / 4 & \textcolor{darkblue}{27.31} & \textcolor{darkblue}{0.766} & {28.25} & {0.782}\\
        SRResNet-Ours & 4 / 4 & \textcolor{darkred}{27.35} & \textcolor{darkred}{0.768} & \textcolor{darkred}{28.80} & \textcolor{darkred}{0.812}\\
        \bottomrule
    \end{tabular}
    }
    }
    \caption{Performance comparison with PTQ methods using EDSR and SRResNet with a scale factor of 4 on larger datasets.}
    \label{tab:largedataset}
\end{table}
\subsection{Experimental Setup}
In our experiments, we follow previous methods \cite{tu2023toward, hong2024adabm} and build the calibration set by randomly sampling $100$ low-resolution images from the DIV2K \cite{agustsson2017ntire} training dataset, without including ground truth images. 
Following the setting in \cite{tu2023toward, hong2024adabm}, the test sets include Set5 \cite{bevilacqua2012low}, Set14 \cite{zeyde2012single}, BSD100 \cite{martin2001database}, and Urban100 \cite{huang2015single}.
We also consider larger datasets, including Test2K and Test4K \cite{kong2021classsr}, which are generated by downsampling the images in the DIV8K dataset \cite{gu2019div8k}. 
We evaluate our method on representative SR networks, including EDSR \cite{lim2017enhanced}, RDN \cite{zhang2018residual}, and SRResNet \cite{ledig2017photo}. 
For EDSR network, we chose the model configuration that consists of $16$ residual blocks with $64$-channel dimensions.
For evaluation, we calculate Peak Signal-to-Noise Ratio (PSNR) and Structural Similarity Index Measure (SSIM) \cite{wang2004image} between the quantized image and the corresponding high-resolution image on the Y channel.

In the calibration phase, we conduct calibration for one epoch using a batch size of \(N=16\). In the first batch, we employ min-max \cite{jacob2018quantization} to establish initial quantization ranges for weights and activations.
The breakpoint is set by taking the $99$th percentile values of the activations for each layer. 
In subsequent batches, the exponential moving average (EMA) hyperparameter \(\beta\) is set to a fixed value of $0.9$. 
In the fine-tuning phase, we utilize Adam optimizer \cite{diederik2014adam} to optimize the clipping ranges of weights, activations, and breakpoints over $10$ epochs with a batch size of $2$. 
We set the hyperparameter \(\lambda = 5\).
The initial learning rate is set to $0.001$ and decays by a factor of $0.9$.

For comparison with PTQ methods, we follow \cite{tu2023toward} and quantize all layers in the models for the Set5, Set14, BSD100, and Urban100 datasets, with the first and last layers quantized at $8$-bit precision. For the Test2K and Test4K datasets, we follow \cite{hong2024adabm} and exclude the first and last layers from quatization. 
For comparison with QAT baselines, we follow \cite{hong2022cadyq, li2020pams, tian2023cabm} and also ignore the first and last layers in the quantization process.

\begin{figure*}[t]
    \centering
    \includegraphics[width=0.92\textwidth, trim=0 11.5cm 0 0, clip]{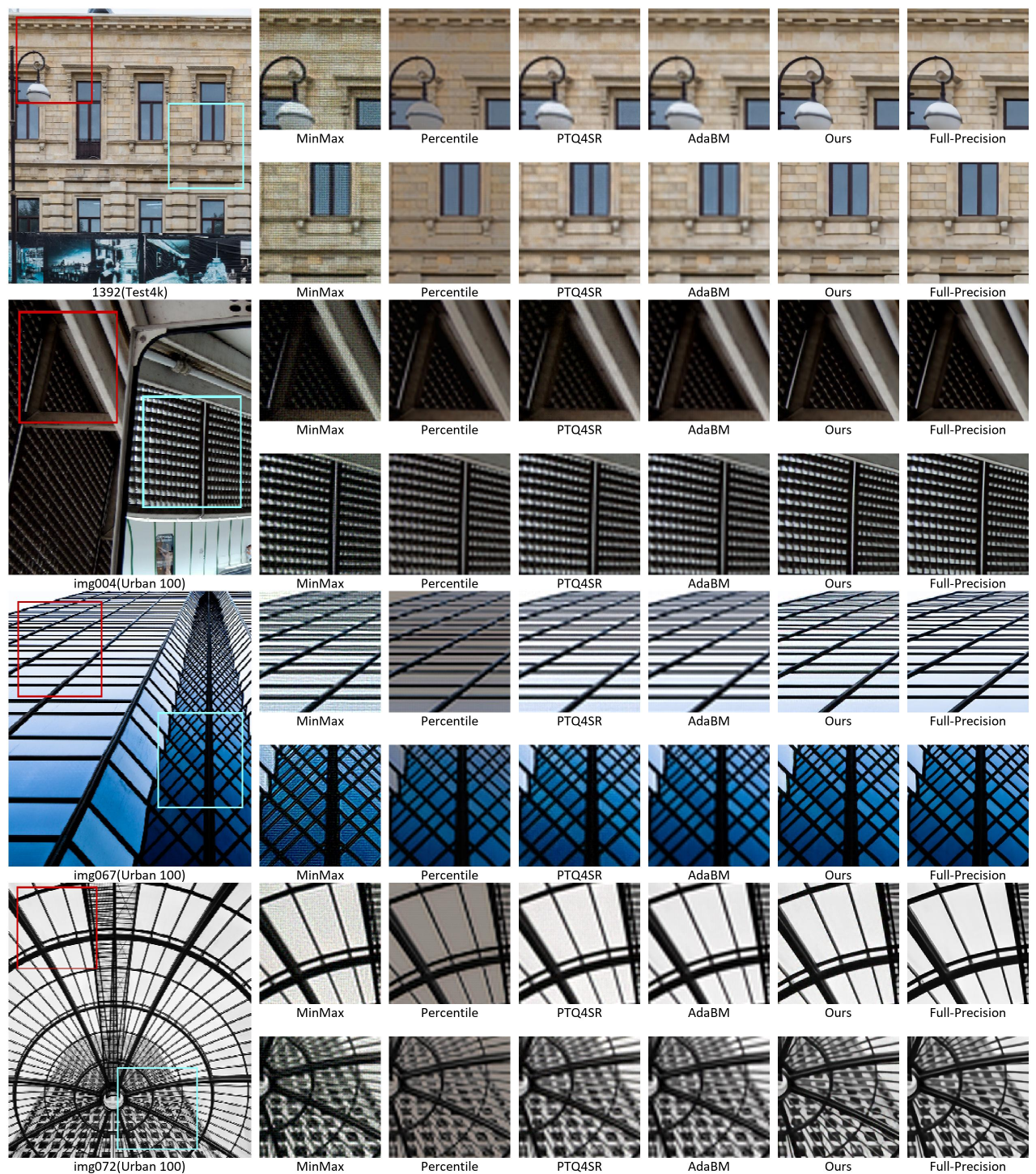}
    \caption{Visual comparison between different PTQ methods using RDN network under W4A4 setting. While baseline approaches suffer from different artifacts, our method effectively preserves the fine details across various scenarios.}
    \label{fig:visulize}
\end{figure*}

\begin{table*}[!tb]
    \centering
    \scalebox{0.85}{    
    \begin{tabular}{lcccccccccccc}
        \toprule
        \multirow{2}*{Method} & \multirow{2}*{QAT} & \multirow{2}*{GT} & \multirow{2}*{W / A} & \multirow{2}*{Process Time}
        & \multicolumn{2}{c}{Set5} & \multicolumn{2}{c}{Set14} & \multicolumn{2}{c}{BSD100} & \multicolumn{2}{c}{Urban100} \\
        \cmidrule(lr){6-7}\cmidrule(lr){8-9}\cmidrule(lr){10-11}\cmidrule(lr){12-13}
        & & & &  & PSNR&SSIM & PSNR&SSIM & PSNR&SSIM & PSNR&SSIM \\
        \midrule
        EDSR & - & $\checkmark$ & 32/32 & - & 32.10 & 0.894 & 28.58 & 0.781 & 27.56 & 0.736 & 26.04 & 0.785 \\
        \midrule
        EDSR-PAMS & $\checkmark$ & $\checkmark$ & 4/4 & 75$\times$  & 31.59 & 0.885 & 28.20 & 0.773 & 27.32 & 0.728 & 25.32 & 0.762 \\
        EDSR-DAQ & $\checkmark$ & $\checkmark$ & 4/4 & 185$\times$  & 31.85& 0.887 & 28.38 & 0.776 & 27.42 & 0.732 & 25.73 & 0.772 \\
        EDSR-DDTB & $\checkmark$ & $\checkmark$ & 4/4 & 125$\times$ & \textcolor{darkblue}{31.85} & \textcolor{darkblue}{0.889} & 28.39 & 0.777 & 27.44 & \textcolor{darkblue}{0.732} & 25.69 & \textcolor{darkblue}{0.774} \\
        EDSR-ODM & $\checkmark$ & $\checkmark$ & 4/4 & 120$\times$  & \textcolor{darkred}{32.00} & \textcolor{darkred}{0.891} & \textcolor{darkred}{28.47} & \textcolor{darkred}{0.779} & \textcolor{darkred}{27.51} & \textcolor{darkred}{0.735} & \textcolor{darkred}{25.80} & \textcolor{darkred}{0.778} \\
        EDSR-Ours & $\times$ & $\times$ & 4/4 & 1$\times$ & 31.79 & 0.885 &  \textcolor{darkblue}{28.40} & \textcolor{darkblue}{0.778} &  \textcolor{darkblue}{27.45} & 0.731 & \textcolor{darkblue}{25.75} & {0.773} \\
        \bottomrule
    \end{tabular}
    }
    \caption{Comparison with QAT methods using EDSR network. Process time was measured on an NVIDIA GeForce RTX 2080Ti GPU.}
    \label{tab:qat}
\end{table*}

\subsection{Comparison with Post-Training Quantization}
We compare the proposed method with existing PTQ approaches, including MSE \cite{choi2018bridging}, Percentile \cite{li2019fully}, MinMax \cite{jacob2018quantization}, PTQ4SR \cite{tu2023toward}, and AdaBM \cite{hong2024adabm},
across the EDSR \cite{lim2017enhanced}, RDN \cite{zhang2018residual} and SRResNet \cite{ledig2017photo} networks. 
The quantitative results for a scale factor of $4$ are presented in Table \ref{tab:ptq}, while results for SRResNet are included in the supplementary appendix.
As shown, our method consistently delivers the top results on every dataset.
Notably, our approach demonstrates a greater advantage over existing approaches on detail-rich datasets and under challenging quantization settings.
For instance, when quantizing the RDN model under the W4A4 setting on the Urban100 dataset, our method achieves a substantial PSNR improvement of $2.55$ dB over the suboptimal method. 
Additionally, we observe that across all settings, the MinMax method significantly outperforms the Percentile method, highlighting the importance of preserving outliers in SR tasks.
Table \ref{tab:largedataset} presents the comparison results on the larger Test2K and Test4K datasets. As shown, our method consistently outperforms existing PTQ approaches, particularly in challenging settings such as W4A4, further demonstrating its robustness and effectiveness in low-bit quantization scenarios.

\subsection{Comparison with Quantization-aware Training}
To further demonstrate our method’s effectiveness, we benchmark it against existing QAT baselines, including PAMS \cite{li2020pams}, DAQ \cite{hong2022daq}, DDTB \cite{zhong2022dynamic} and ODM \cite{hong2024overcoming}, on the EDSR \cite{lim2017enhanced} networks.
In practice, QAT methods typically require several hours for quantization due to the need for retraining model parameters. In contrast, our proposed method completes the quantization process in less than 2 minutes, significantly improving efficiency.
Table \ref{tab:qat} presents the comparison with a scale factor of 4. 
As shown, compared to QAT methods, our approach achieves at least a $75\times$ speedup while achieving comparable performance without requiring ground truth supervision.

\subsection{Qualitative Analysis}
To provide a more intuitive assessment of performance, we present the SR results for each method in Figure \ref{fig:visulize}. 
As illustrated, images produced by MinMax contain numerous noise artifacts, as preserving outliers in MinMax restricts the expressiveness of normal activations and distorts the image distribution with noise.
In contrast, the images generated by Percentile exhibit significant color distortion, highlighting the importance of outliers in maintaining color fidelity.
While PTQ4SR and AdaBM mitigate noise and color shift issues to some extent, they still introduce blurring in detail-rich areas, particularly in dense regions of images img$004$ and img$072$, leading to a noticeable decline in visual fidelity. By examining intricate textures and flat regions side by side, we find that our method effectively preserves fine details while avoiding noise and color distortion, demonstrating superior SR quality.

\subsection{Ablation Study}
To assess the efficacy of our proposed piecewise linear quantizer (PLQ) and sensitivity-aware finetuning (SAFT) strategies, we conduct an ablation study using MinMax as the baseline model and assess the impact of these two components.
The results are presented in Table \ref{tab:abla}. 
As shown, combining both PLQ and SAFT (5th row) achieves the best performance. Compared to the baseline (1st row), applying PLQ alone (2nd row) leads to a significant PSNR improvement, with gains of $3.67$ dB, $2.67$ dB, $2.46$ dB, and $2.00$ dB on Set5, Set14, BSD100, and Urban100, respectively. 
Additionally, we observe that vanilla finetuning (VFT, 3rd row) performs worse than SAFT (4th row), which highlights the effectiveness of our proposed sensitivity-aware loss in improving quantization performance.

\begin{table}[!tb]
    \centering
    \scalebox{1.0}{    \resizebox{\columnwidth}{!}{ 
    \begin{tabular}{ccccccc}
        \toprule
         \multirow{2}{*}{PLQ} & \multirow{2}{*}{SAFT} & \multirow{2}{*}{VFT} & {Set5} & {Set14} & {BSD100} & {Urban100} \\
         &  &  &  PSNR / SSIM & PSNR / SSIM & PSNR / SSIM & PSNR / SSIM \\
        \midrule
        $\times$ & $\times$ & $\times$ & 26.83 / 0.624 & 25.04 / 0.546 & 24.57 /  0.503 & 23.12 / 0.536 \\
        $\checkmark$ & $\times$ & $\times$ & 30.50 / 0.865 & 27.71 / 0.755 & 27.03 / 0.715 & 25.12 / 0.751 \\
        $\times$ & -- & $\checkmark$  & 29.45 / 0.770 & 26.95 / 0.677 & 26.27 /  0.632 & 24.40 / 0.654 \\
        $\times$ & $\checkmark$ & -- & 29.87 / 0.809 & 27.24 / 0.709 & 26.55 / 0.666 & 24.57 / 0.689 \\
        $\checkmark$ & $\checkmark$ & -- & \textcolor{darkred}{31.54} / \textcolor{darkred}{0.879} & \textcolor{darkred}{28.26} / \textcolor{darkred}{0.769} & \textcolor{darkred}{27.36} / \textcolor{darkred}{0.726} & \textcolor{darkred}{25.61} / \textcolor{darkred}{0.765} \\
        \bottomrule
    \end{tabular}
    }
    }
    \caption{Ablation study on EDSR network under W4A4 setting. PLQ, SAFT, VFT denotes Piecewise Linear Quantizer, Sensitivity-Aware Finetuning, Vanilla Finetuning, respectively.}
    \label{tab:abla}
\end{table}

\subsection{Resource Analysis}
To validate the efficiency of our method, we compare its processing time, latency (a single forward-pass inference time), and storage with exiting PTQ baselines. As shown in Table \ref{tab:hardware}, our method reduces processing time compared to PTQ4SR, while remaining comparable to AdaBM. In terms of latency, our approach performs similarly to PTQ4SR and is faster than AdaBM. Additionally, all methods maintain the same storage size. These results demonstrate that our performance improvements are achieved without increasing resource demands. 

 \begin{table}[!tb]
    \centering
    \scalebox{0.46}{
    \resizebox{\textwidth}{!}{
    \begin{tabular}{lccc}
        \toprule
        & Process Time & Latency &
        Storage size \\
        \midrule
        EDSR-PTQ4SR & 126 sec & 133 ms  & 229.517K\\
        EDSR-AdaBM & 72 sec & 143 ms & 229.517K\\
        EDSR-Ours & {73 sec} & 135 ms & 229.517K\\
        \bottomrule
    \end{tabular}
    }
    }
    \caption{Efficiency comparison with PTQ methods using a scale factor of 4. Processing time and latency (fake quantization) are measured on an NVIDIA 2080Ti GPU.}
    \label{tab:hardware}
\end{table}

\section{Conclusion}
This paper introduces an outlier-aware post-training quantization method for image super-resolution tasks. 
According to our empirical analysis on activation distribution, we observe that outliers in activations are both ubiquitous and impactful.
We then conduct comparison experiments to investigate the impact of outliers and uncover that the outliers are strongly correlated with image color information.
Specifically, simply removing outliers in activations will cause noticeable color distortion and considerable performance degradation.
However, retaining them will reduce bit occupancy reserved for normal activations. 
To strike a balance between preserving outliers and maintaining quantization effectiveness on normal activations, we divide the activation distribution into two non-overlapping regions and apply uniform quantization to each region independently.
Additionally, motivated by our finding that different network layers exhibit varying sensitivities to quantization, we design a sensitivity-aware loss function to make the model focus more on highly sensitive layers.
We then conduct extensive experiments to demonstrate the effectiveness of our method, comparing it against both PTQ and QAT approaches across various datasets and model architectures.

\clearpage
{
    \small
    \bibliographystyle{ieeenat_fullname}
    \bibliography{main}
}

\end{document}